\documentclass[pdflatex,sn-mathphys-num]{sn-jnl}% Math and Physical Sciences Numbered Reference Style 
\usepackage{graphicx}%
\usepackage{multirow}%
\usepackage{amsmath,amssymb,amsfonts}%
\usepackage{amsthm}%
\usepackage{mathrsfs}%
\usepackage[title]{appendix}%
\usepackage{xcolor}%
\usepackage{textcomp}%
\usepackage{manyfoot}%
\usepackage{booktabs}%
\usepackage{algorithm}%
\usepackage{algorithmicx}%
\usepackage{algpseudocode}%
\usepackage{listings}%
\usepackage{subfig}
\theoremstyle{thmstyleone}%
\theoremstyle{thmstyletwo}%
\usepackage{threeparttable}
\theoremstyle{thmstylethree}%

\begin{document}

\title[scAdaDrug]{Towards generalization of drug response prediction to single cells and patients utilizing importance-aware multi-source domain transfer learning}

\author[1]{\fnm{Hui} \sur{Liu}}%\email{hliu@njtech.edu.cn}
\author[1]{\fnm{Wei} \sur{Duan}}

%\author[1]{\fnm{Xing} \sur{Fang}}
%\author[1]{\fnm{Shiye} \sur{Tian}}
\author*[2]{\fnm{Judong} \sur{Luo}}\email{judongluo@tongji.edu.cn}

\affil[1]{\orgdiv{College of Computer and Information Engineering}, \orgname{Nanjing Tech University}, \orgaddress{\city{Nanjing}, \postcode{211800 }, \state{Jiangsu}, \country{China}}}

\affil[2]{\orgdiv{Department of Radiotherapy}, \orgname{Tongji Hospital, School of Medicine, Tongji University}, \orgaddress{\city{Shanghai}, \postcode{200065},  \country{China}}}


\abstract{The advancement of single-cell sequencing technology has promoted the generation of a large amount of single-cell transcriptional profiles, providing unprecedented opportunities to identify drug-resistant cell subpopulations within a tumor. However, few studies have focused on drug response prediction at single-cell level, and their performance remains suboptimal. This paper proposed scAdaDrug, a novel multi-source domain adaptation model powered by adaptive importance-aware representation learning to predict drug response of individual cells. We used a shared encoder to extract domain-invariant features related to drug response from multiple source domains by utilizing adversarial domain adaptation. Particularly, we introduced a plug-and-play module to generate importance-aware and mutually independent weights, which could adaptively modulate the latent representation of each sample in element-wise manner between source and target domains. Extensive experimental results showed that our model achieved state-of-the-art performance in predicting drug response on multiple independent datasets, including single-cell datasets derived from both cell lines and patient-derived xenografts (PDX) models, as well as clinical tumor patient cohorts. Moreover, the ablation experiments demonstrated our model effectively captured the underlying patterns determining drug response from multiple source domains.}




\maketitle

\section{Introduction}
The intrinsic and acquired resistance to drugs are the main causes of failure in clinical cancer therapy~\cite{camidge2014acquired,vasan2019view,rood2022impact}. Despite the fact that routine drug treatment can effectively eliminate the majority of malignant cells, a small subset of tumor cells would survive and continue to proliferate, ultimately leading to tumor recurrence and progression~\cite{restifo2016acquired,aissa2021single}. In recent years, the advent of single-cell RNA-sequencing (scRNA-seq) technology has generated a wealth of single-cell transcriptional data~\cite{franzen2019panglaodb,sun2021tisch}, providing valuable opportunity to identify drug-resistant cell subpopulations~\cite{gambardella2022single,liu2024drmref}. However, there remains a significant gap in our knowledge regarding drug response at the single-cell level, pressing a challenging but urgent task to predict the drug sensitivity of individual cells.

Domain adaptation, which enables the knowledge transfer from source domains to target domains, has achieved remarkable success in the field of computer vision~\cite{ganin2015unsupervised,zhao2018adversarial,saito2018maximum}. In recent years, domain adaptation has successfully been applied to the field of bioinformatics, such as domain transfer from cancer cell lines to individual cells or patients~\cite{suphavilai2021predicting,zhao2021deconvolution}. For instance, He et al. proposed the CODE-AE model~\cite{he2022context}, which utilized domain separation network~\cite{bousmalis2016domain} to extract domain-invariant features between cell lines and patients. This model was trained on cell line drug sensitivity data and then used to predict the drug response for tumor patients. Chawla et al. developed Precily~\cite{chawla2022gene} that integrated signaling pathways and drug features to predict drug responses in vitro and in vivo. Inspired by the Fader network~\cite{lample2017fader} and the compositional perturbation autoencoder (CPA)~\cite{lotfollahi2023predicting}, Hetzel et al. proposed ChemCPA~\cite{hetzel2022predicting} to combine chemical property and drug-induced expression profiles to infer transcriptional responses to unseen drug perturbations. Other models, such as DeepCE~\cite{pham2021deep} and GEARS~\cite{roohani2022gears}, have been developed to predict drug-induced transcriptional profiles.

Due to the limited availability of single-cell drug response data, a few studies have leveraged domain adaptation between bulk RNA-seq (source domain) and scRNA-seq (target domain) data to predict drug sensitivity of individual cells. For example, Chen et al. proposed scDEAL~\cite{chen2022deep} to align the bulk RNA-seq and scRNA-seq features by minimizing the maximum mean discrepancy (MMD)~\cite{gretton2012kernel} in the latent space, so that the classifier trained on bulk drug sensitivities can be transferred to predict single-cell drug sensitivity. Zheng et al. developed SCAD~\cite{zheng2023enabling}, which was based on adversarial domain adaptation, to learn drug-gene signatures from the GDSC dataset~\cite{yang2012genomics} for inferring drug sensitivity in single cells. However, all previous methods employed only one source domain, which limited their ability to learn the essential feature from diverse types of cells determining drug treatment. Therefore, multi-source domain adaptation (MDA) has been proposed to transfer knowledge from multiple source domains to target domain~\cite{sun2015survey,sun2011two,duan2012exploiting,schweikert2008empirical}. For instance, Zhao et al. introduced a multi-source domain adaptation (MDDA) model~\cite{zhao2020multi} that minimized the empirical Wasserstein distance between source and target domains, attaining high performance in image classification tasks. Pei et al. proposed a method called multi-adversarial domain adaptation (MADA)~\cite{pei2018multi}, which focused on class-level alignment of different data distributions using multiple domain discriminators. Additionally, Fu et al. proposed PFSA~\cite{fu2021partial} to address the category shift problem where the numbers and types of categories in the source and target domains were inconsistent.

Inspired by multi-source domain adaptation, we proposed scAdaDrug to predict the drug response of single cells. In this study, we consider the bulk RNA-seq data of multiple cell lines with drug response labels to constitute the multiple source domains, with the scRNA-seq data serving as the target domain. Different from existing methods assigning uniform importance to each source domain, we introduce a plug-and-play module to produce an importance-aware weight vector that captured element-wise relevance between each source domain and target domain. Under the conditional independence constraint, the weight generator enforced the encoder to extract non-redundant features from multiple source domains. Furthermore, through adversarial learning, our model can learn domain-invariant features related to drug response, so that our model can generalize its predictive power to target-domain drug responses.

The main contributions of this study are as follows:
\begin{itemize}
    \item We propose a novel multi-source domain adaptation framework for drug response prediction, transferring predictive capacity from the label-rich cell lines to label-scarce single cells and tumor patients. 
    \item Considering the difference of importance in transferring knowledge from multiple source domains to target domain, we devised a plug-and-play module that can generate an importance-aware weights to capture fine-grained relevance between source and target domains.
    \item To avoid information redundancy among multiple source domains, we imposed conditional independence constraints on the generated weights, thereby allowing to learn causally independent features from multiple source domains and aligned to target domain.
    \item Our extensive experimental results demonstrated that the proposed model achieved state-of-the-art performance in transfer knowledge learned from bulk RNA-seq data to predict the single-cell and patient-level drug responses.
\end{itemize}

\section{Results}\label{Results}
\subsection{Importance-aware multi-source domain adaptation}
\begin{figure*}[htbp]
    \centering
    \centerline{\includegraphics[width=17cm]{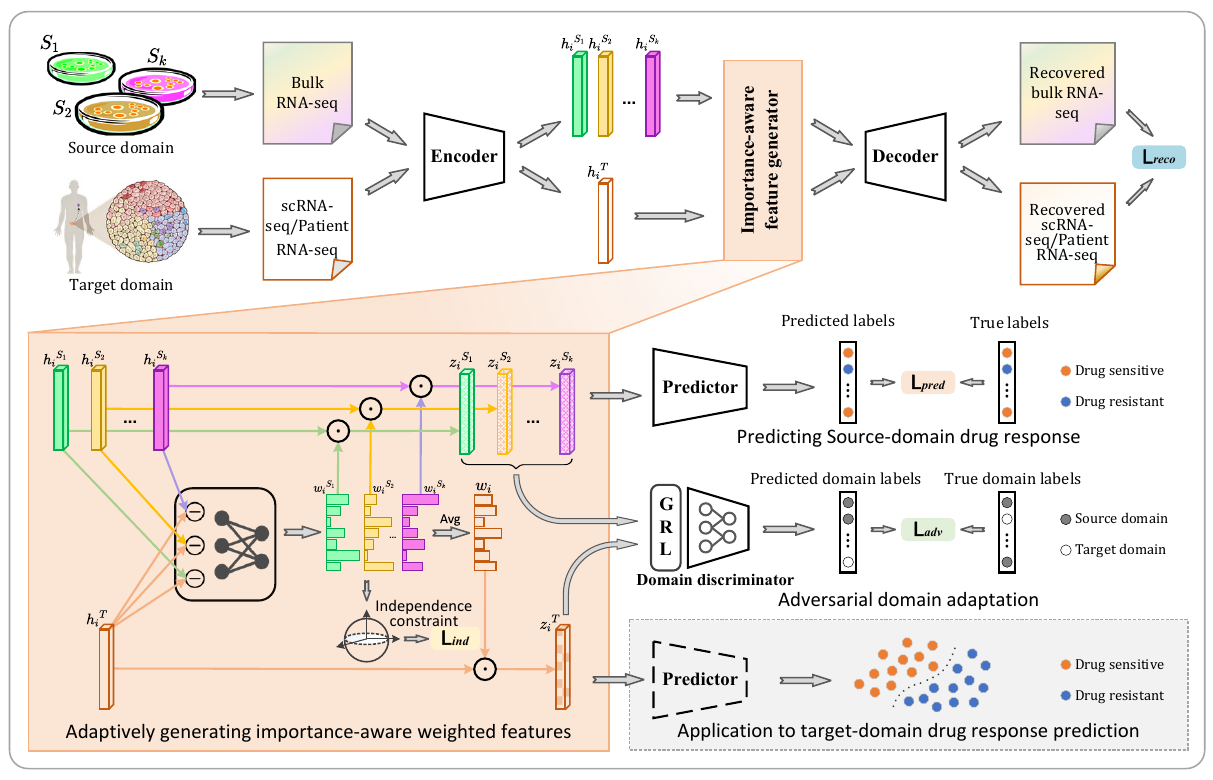}}
    \caption{Illustrative diagram of the proposed scAdaDrug architecture for predicting single-cell drug response. It consists of four components: autoencoder-based feature extractor, adaptive weight generator, adversarial domain discriminator and drug sensitivity predictor.} \label{fig:framework}
\end{figure*}

We proposed a novel multi-source domain adaptation to transfer knowledge learned from bulk drug sensitivity of cell lines to predict single-cell drug response. For simplicity, we refer to the proposed model as scAdaDrug in the following text. In the following context, bulk RNA-seq data from multiple cell lines of a single cancer type were conceptualized as multi-source domains, while scRNA-seq data from the same cancer type were designated as the target domain. Each cell line in the source domains has been assigned response label regarding a specific drug (sensitive labeled as 1 and resistant as 0). The scAdaDrug model consists of four components: autoencoder-based feature extractor, importance-aware weight generator, adversarial domain discriminator and drug response predictor (Fig.~\ref{fig:framework}). A shared autoencoder is used to extract the features for both source and target domains. Specifically, we introduce an adaptive weight generator to produce an importance-aware weights that are able to capture element-wise relevance between source and target domains. Meanwhile, we required that the generated weights were conditionally independent so as to eliminate information redundancy between multiple domain sources. Therefore, the adaptively generated weights allow for fine-grained and non-redundant knowledge transfer from source domains to target domain. Furthermore, we incorporated adversarial learning to learn domain-level invariant features related to drug response. Our extensive experiments confirmed that scAdaDrug is able to generalize well across different target domains, thereby improving its predictive performance of drug response for both individual cells and patients.

\subsection{Performance comparison on single-cell drug response of cell lines}\label{subsec2}
Our primary objective is to predict drug responses at the single-cell level, which would facilitate to identify a subgroup of drug-resistant tumor cells. The bulk RNA-seq data of cell lines (source domains) was obtained from the GDSC database, while the scRNA-seq data (target domain) were obtained from GEO (accession numbers: GSE149215 and GSE108383). The first scRNA-seq dataset comprised transcriptional profiles and drug response labels of 1,393 single cells from PC9 cell line after Etoposide treatment~\cite{aissa2021single}. The second scRNA-seq dataset included transcriptional profiles and drug response labels of 108 single cells from the A375 cell line and 197 single cells from 451Lu cell line after PLX4720 treatment~\cite{ho2018single}. The details of the source and target domain datasets with common drugs were listed in TABLE~S1.

\begin{figure*}[htbp]
    \centering
    \centerline{\includegraphics[width=17.5cm]{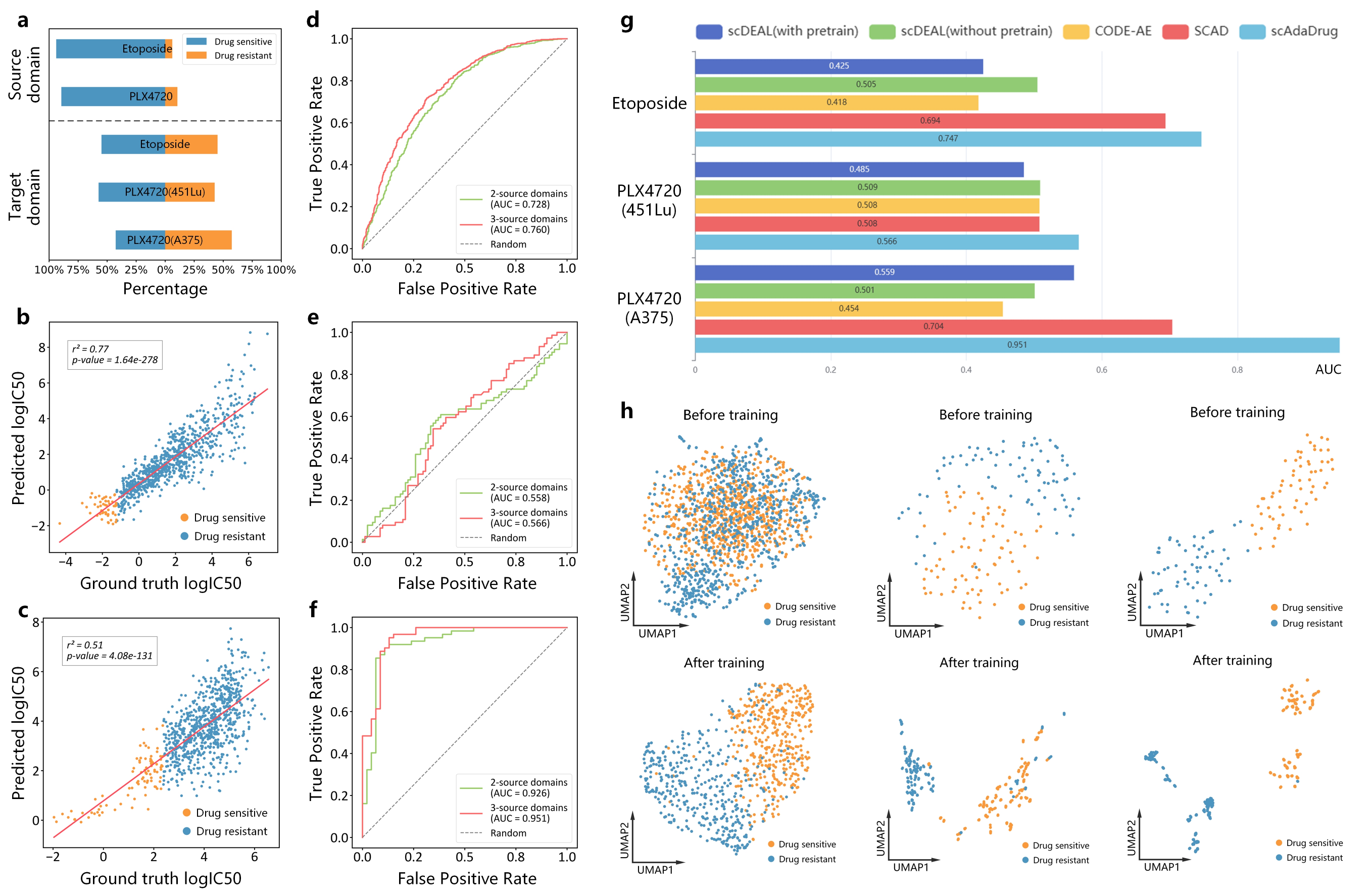}}
    \caption{Performance evaluation on single-cell drug response datasets of cell lines. \textbf{a} The imbalanced number of drug-sensitive and drug-resistant cells in both source and target domain. \textbf{b-c} The scatter plots between the predicted and true log(IC50) values of cell lines treated by Etoposide, PLX4720, respectively. \textbf{d} ROC curve of scAdaDrug for predicting single-cell drug response of PC9 cell line treated by  Etoposide; \textbf{e-f} ROC curves of scAdaDrug for predicting single-cell drug response of PLX4720-treated 451Lu and A375 cell lines, respectively. \textbf{g} Performance comparison of scAdaDrug to four competing methods on three single-cell datasets. \textbf{h} UMAP feature plots of single-cell transcriptional profiles and corresponding embeddings output by the trained encoder for Etoposide-treated PC9, PLX4720-treated 451Lu and PLX4720-treated A375 cell lines, respectively.} 
    \label{fig2_exp1_SCAD}
\end{figure*}

Meanwhile, to explore whether the number of input genes affects performance, we selected a set of highly variable genes for performance comparison. For this purpose, we selected a set of highly variable genes that exhibited most significant difference in expression levels shared in source and target domains. First, the top 4,000 highly variable genes in scRNA-seq were identified by the function of highly\_variable\_genes in Scanpy, which is a popular method for denoising and feature selection strategy prior to downstream analysis on scRNA-seq data~\cite{wolf2018scanpy}. Also, we extracted the genes involved in 2,128 protein–protein interactions from a prior knowledge graph that has compiled most informative genes relevant to drug sensitivity~\cite{manica2019toward}. As a result, we identified the overlapped genes to serve as model inputs. Due to the significant imbalance in the number of sensitive and resistant cell lines in the source domains (Fig.~\ref{fig2_exp1_SCAD}a), we tried both weighted sampling and SMOTE sampling to alleviate the influence of class imbalance on model decision boundary. Meanwhile, we sought to investigate the impact of varying number of source domains on the predictive performance in target domain. So, we trained two model variants with two and three source domains for performance evaluation. The experimental results demonstrated that  three-source-domain model achieves superior performance than two-source-domain model  across all three datasets. (Fig.~\ref{fig2_exp1_SCAD}d-f). 

We conducted performance comparison with four existing methods, including CODE-AE~\cite{he2022context}, scDEAL~\cite{chen2022deep}, and SCAD~\cite{zheng2023enabling}. We evaluated two variants of scDEAL, one with pretraining and one without. All the competing methods were trained and evaluated on the same dataset. As a result, our method achieved higher performance than all competing methods in predicting the single-cell response regarding three drugs. Although our method exhibited suboptimal performance in PLX4720-treated 451Lu cells, it still outperformed these competing methods (Fig.~\ref{fig2_exp1_SCAD}g). %The suboptimal performance observed in PLX4720-treated 451Lu cells can be attributed to the inherent characteristics in transcriptomic profiles of specific cell type~\cite{ho2018single}. 

Moreover, since the benchmark datasets were released by SCAD, we conducted a more in-depth comparison with SCAD to further verify the performance of our proposed method. As expected, the experimental results demonstrate that our model consistently outperformed SCAD across the drug-induced scRNA-seq datasets (TABLE~\ref{tab:scAdaDrug2SCAD_AUROC}; TABLE S2), regardless of the upsampling strategy and number of source domains. In particular, when utilizing three source domains with SMOTE upsampling, our model achieved the best performance. The reason may lie in that the high dimensionality of transcriptional profiles often lead to model overfitting, while SMOTE upsampling mitigates this problem by creating new minority class samples, thereby improving the model's ability to generate smoother decision boundaries~\cite{blagus2013smote}. Upon closer examination, it can be seen that, given the same upsampling algorithm, the performance was improved when built on three source domains compared to two. We also found that using highly variable genes as input can bring about about 2\% performance improvement. We speculate that the use of highly variable genes facilitated the model to capture biologically discriminative feature more effectively. We applied Uniform Manifold Approximation and Projection (UMAP) dimensionality reduction~\cite{mcinnes2018umap} for visualization of the cells with similar gene expression signatures, and generated the UMAP plots of the single-cell transcriptional profiles and the corresponding embeddings output by the trained encoder (Fig.~\ref{fig2_exp1_SCAD}h). It was observed that the original transcriptional profiles of drug-sensitive and drug-resistant cells were mixed together, yet the learned embeddings separated significantly in the latent space. This indicated that our model effectively captures the discriminative features determining drug resistance and sensitivity of individual cells.

\renewcommand{\arraystretch}{1.2}
\begin{table*}[h]
\centering
\caption{AUROC values achieved by scAdaDrug and SCAD in predicting single-cell drug sensitivities}\label{tab:scAdaDrug2SCAD_AUROC}
\resizebox{1\textwidth}{!}{%
\begin{threeparttable}
\begin{tabular}{c|cccc|cccc|cccc} \hline
Method & \multicolumn{4}{c|}{scAdaDrug (2 source domains)} & \multicolumn{4}{c|}{scAdaDrug (3 source domains)} & \multicolumn{4}{c}{SCAD} \\ \hline
Drugs & Weight-HVG & Smote-HVG &Weight-ALL &Smote-ALL & Weight-HVG & Smote-HVG & Weight-ALL & Smote-ALL & Weight-HVG & Smote-HVG & Weight-ALL & Smote-ALL \\ \hline
Etoposide & 0.710 & 0.739 & 0.727 & 0.719 & 0.727 & \textbf{0.747} & 0.731 & 0.725 & 0.583 & 0.671 & 0.669 & 0.694\\
PLX4720(451Lu) & 0.512 & 0.558 & 0.550 & 0.551 & 0.532 & 0.566 & 0.569 & \textbf{0.570} & 0.508 & 0.504 & 0.381 & 0.381\\
PLX4720(A375) & 0.941 & 0.951 & 0.824 & 0.917 & 0.948 & \textbf{0.951} & 0.835 &0.918 & 0.780 & 0.937 & 0.694 & 0.825\\
Average & 0.724 & 0.746 & 0.700 & 0.729 & 0.737 & \textbf{0.754} & 0.712 & 0.738 & 0.624 & 0.704 & 0.581 & 0.633\\ \hline
\end{tabular}
\begin{tablenotes}
        \item Weight: weight upsampling; Smote: smote upsampling; ALL: all genes shared by source and target domains; HVG: highly variable genes shared by source and target domains.
\end{tablenotes}
\end{threeparttable}}
\end{table*}
\renewcommand{\arraystretch}{1.0}

\subsection{Predicting single-cell drug response of PDX samples}
\begin{figure*}[htbp]
    \centering
    \centerline{\includegraphics[width=16cm]{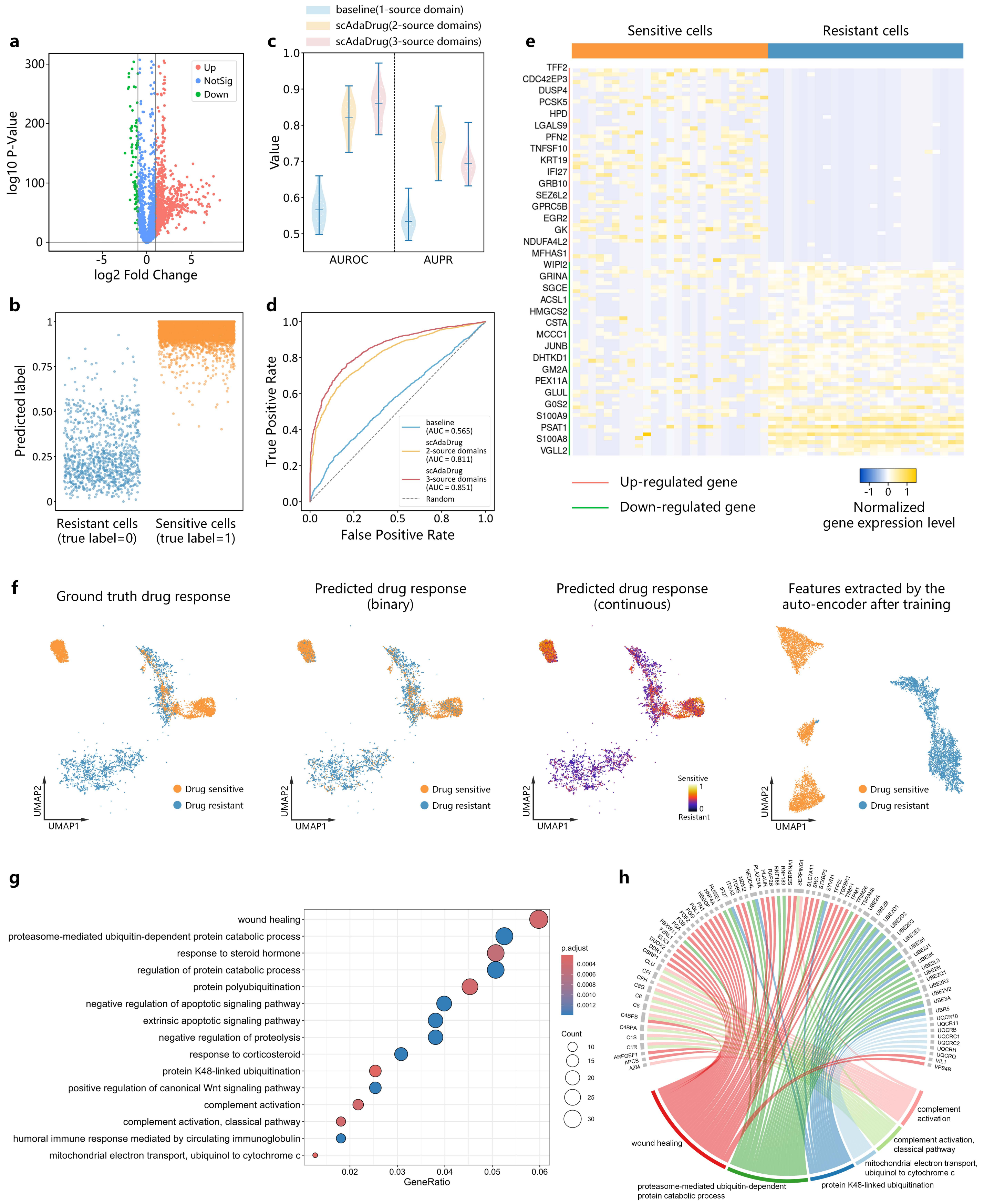}}
    \caption{Performance evaluation in predicting single-cell drug response of PDX samples of hepatocellular carcinoma (GSE175716). \textbf{a} Volcano plot of differentially expressed genes between sensitive and resistant cells upon sorafenib treatments. \textbf{b} The scAdaDrug predicts labels for drug-resistant and drug-sensitive cells from PDX samples. \textbf{c-d} Violin plots and ROC curves achieved by baseline model and scAdaDrug (2- and 3-source domains) in predicting single-cell drug response. \textbf{e} Normalized expression profiles of top-ranked up-regulated and down-regulated genes. \textbf{f} UMAP  plots of the transcriptional profiles with the individual cells colored by actual response labels, binarized predicted labels, continuous predicted probabilities, as well as the learned embeddings of individual cells colored by actual response labels.\textbf{g-h}Enrichment analysis of differentially expressed genes. \textbf{g-h} GO enrichment analysis using the top 600 up-regulated genes in the drug-resistant cells. } \label{fig3_exp2_sorafenib}
\end{figure*}

To assess the generalizability of scAdaDrug, we evaluated its performance on a scRNA-seq dataset derived from patient-derived xenografts (PDX) samples of hepatocellular carcinoma ~\cite{guan2021activation}. The dataset was obtained from GEO (accession number GSE175716) and comprised scRNA-seq data of 6,823 cells, along with their responses to sorafenib. We selected the liver cancer-related cell lines from the GDSC database as the source domains, with this scRNA-seq dataset as target domain. Inspired by the aforementioned experimental results that highly variable genes enhanced performance, we chosen differentially expressed genes (DEGs) between resistant and sensitive cells as the input of our model. Specifically, we identified 3,448 up-regulated or down-regulated genes ($p$-value$>$0.05, log2 fold change$>$2) from the scRNA-seq data (Fig.~\ref{fig3_exp2_sorafenib}a,e), and subsequently selected out the corresponding set of DEGs from the source domains to construct the input of our model. Meanwhile, due to the significant imbalance between sensitive and resistant cells (5,865 vs 958), we applied SMOTE sampling to avoid biased decision boundary induced by class imbalance.

To benchmark our model performance, we built a single-source domain adaptation model as baseline. For this purpose, we removed the adaptive weight generator, which was designed for multi-source domain adaptation, while keeping the other components of scAdaDrug unchanged. Without the adaptively generated weights, the encoder-extracted features were directly applied for adversarial domain adaptation. When compared to the baseline model, we also considered the performance difference resulted from varying number of source domains. To ensure robustness, we conducted 20 independent training and evaluation cycles with random parameter initialization (no fixed random seed). Performance was measured using AUROC and AUPR, with results visualized through violin plots (Fig.~\ref{fig3_exp2_sorafenib}c-d). The results showed that scAdaDrug significantly outperformed the baseline, achieving  improvements of approximately 25\%-30\% in AUROC and 15\%-20\% in AUPR. The notable enhancements confirmed the effectiveness of the multi-source domain model with adaptive weighted feature. Furthermore, the 3-source domain model achieved an impressive average AUROC of 0.859, surpassing the 2-source domains model, in predicting the single-cell drug sensitivity dataset derived from PDX samples. Additionally, we utilized UMAP tool to visualize the original transcriptional profiles of single cells and the corresponding embeddings generated by the trained encoder (Fig.~\ref{fig3_exp2_sorafenib}f). The UMAP plots clearly illustrated the separation of drug-sensitive and drug-resistant single-cell features in the latent space, whereas the transcriptional profiles were mixed. GO enrichment analysis using the top 600 up-regulated genes in the drug-resistant cells revealed that these genes are enriched in the cellular functions related to drug resistance (Fig.~\ref{fig3_exp2_sorafenib}g-h), such as response to steroid hormone~\cite{hooks2024hormonal}, negative regulation of apoptotic signaling pathway\cite{neophytou2021apoptosis,razaghi2018negative}.

\subsection{Identifying drug-resistant cell population of lung cancer patients}
\begin{figure*}[htbp]
    \centering
    \centerline{\includegraphics[width=16cm]{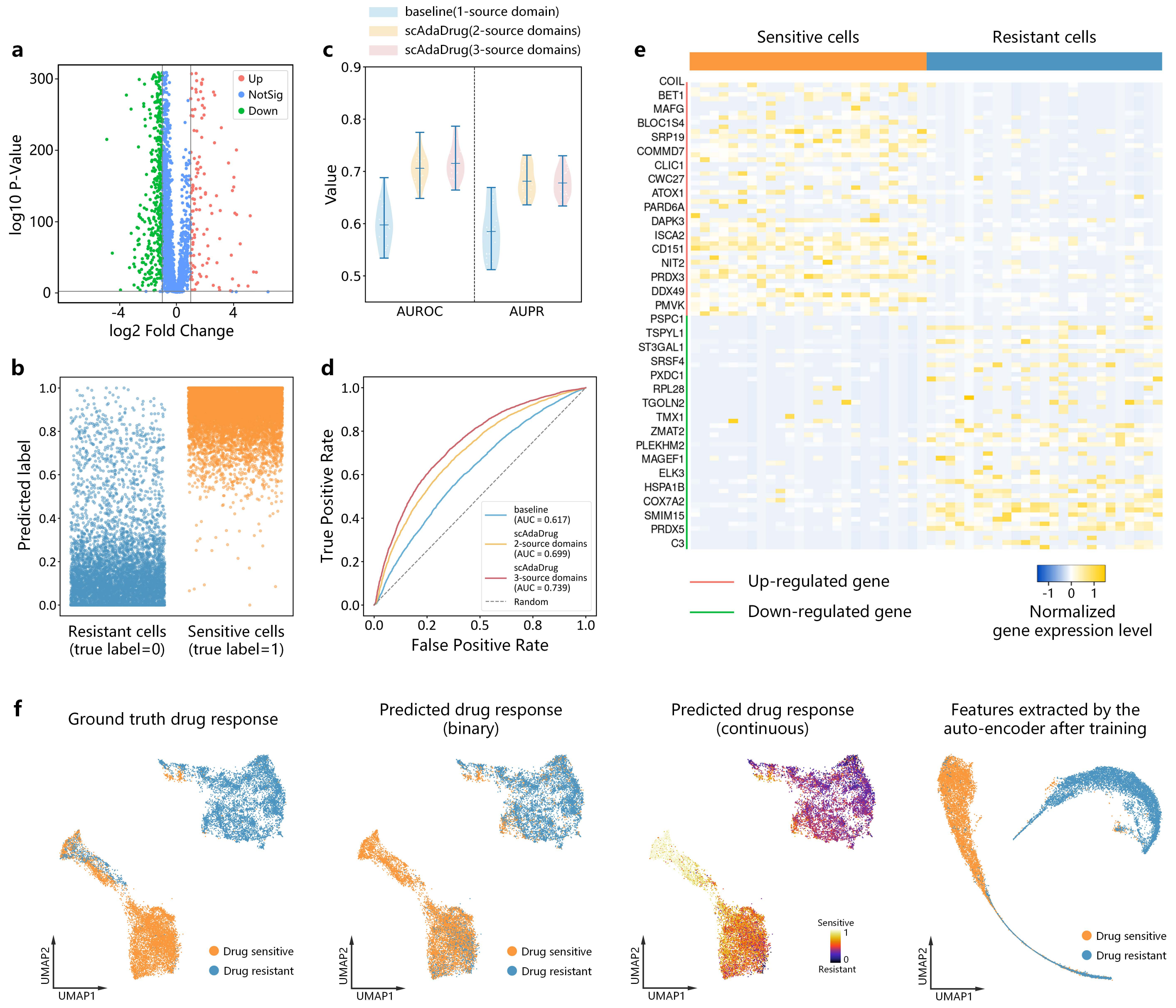}}
    \caption{Performance evaluation in predicting single-cell drug response of patient tissues of lung adenocarcinoma (GSE223779). \textbf{a} Volcano plot of differentially expressed genes between sensitive and resistant cells to crizotinib treatments. \textbf{b} The scAdaDrug predicts labels for drug-resistant and drug-sensitive cells derived from patient tissues. \textbf{c-d} Violin plots and ROC curves achieved by baseline model and scAdaDrug (2- and 3-source domains) in predicting single-cell drug responses. \textbf{e} Normalized expression profiles of top-ranked up-regulated and down-regulated genes. \textbf{f} UMAP plots of the transcriptional profiles with the individual cells colored by actual response labels, binarized predicted labels, continuous predicted probabilities, as well as the learned embeddings of individual cells colored by actual response labels.\textbf{g-h}Enrichment analysis of differentially expressed genes.} 
    \label{fig4_exp3_crizotinib}
\end{figure*}

To test whether our model can discriminate drug-resistant cell population in tumor tissue, we employed the scRNA-seq data of patient tumor tissues from GEO (accession number: GSE223779) as the target domain. This dataset contained the transcriptional profiles of 15,390 single cells derived from lung adenocarcinoma tissues, as well as the single-cell response labels to the drug crizotinib~\cite{kwok2023single}. There were 9,456 cells identified as sensitive, while 5,934 cells exhibited drug resistance. It is worth noting that the actual drug response labels were never used in our model training. Similarly, we leveraged the SMOTE upsampling to tackle class imbalance problem, and took as model input the transcription profiles of DEGs in scRNA-seq data  (Fig.~\ref{fig4_exp3_crizotinib}a,e). Among the top-ranked DEGs, TMX1 and DDX49 have been reported to be involved in drug resistance in lung cancer~\cite{WANGYanan2023,lian2020ddx49}.

The experimental results showed that scAdadrug correctly identified most drug-resistant cells (Fig.~\ref{fig4_exp3_crizotinib}b). We also compared scAdadrug against the baseline model to calibrate its performance (Fig.~\ref{fig4_exp3_crizotinib}c-d). Impressively, scAdadrug outperformed the baseline model by approximately 10\% to 15\% in both AUROC and AUPR metrics, thereby verifying the effectiveness of our multi-source domain adaptation with adaptive weighted features. We again observed that the 3-source domain variant of our model achieved a mean AUROC of 0.715 over 20 randomized repetitions, surpassing the 2-source domain version. For visual presentation, we employed UMAP feature plots to illustrate the transcriptional profiles of individual cells, in which each cell was colored by corresponding drug response labels (Fig.~\ref{fig4_exp3_crizotinib}f). These plots revealed high consistence between the actual and predicted single-cell drug response, highlighting the remarkable predictive accuracy in identifying the set of tumor cells resistant to specific drug treatment. Furthermore, the encoder-derived embeddings distributed separately between resistant and sensitive cells in the latent space. Therefore, scAdaDrug exhibited impressive feature extraction and predictive capabilities in the context of patient-derived single-cell drug sensitivity.

\subsection{Multiple independent datasets validated model generalizability}
To further evaluate the generalizability of scAdaDrug in predicting single-cell drug sensitivity, we utilized five public scRNA-seq datasets to assess its prediction performance. These datasets involved two chemotherapy drugs (Cisplatin~\cite{sharma2018longitudinal,ravasio2020single}, Docetaxel~\cite{schnepp2020single}) and two targeted drugs (Gefitinib~\cite{kong2019concurrent} and Erlotinib~\cite{aissa2021single}), which were administrated to three types of tumors, including oral squamous cell carcinomas, lung cancer and prostate cancer (for details of the datasets see TABLE S3 in Supplementary Data). All datasets have been provided with ground-truth drug response annotations (sensitive or resistant) for individual cells. %From their UMAP plots (Fig.~\ref{fig3_exp2_scDEAL}a-e), it can be seen that the number of sensitive and resistant cells varied greatly across these datasets, as well as the data distributions.

\begin{figure}[htbp]
    \centering
    \begin{minipage}{\textwidth}
        \centering
        \includegraphics[width=0.45\textwidth]{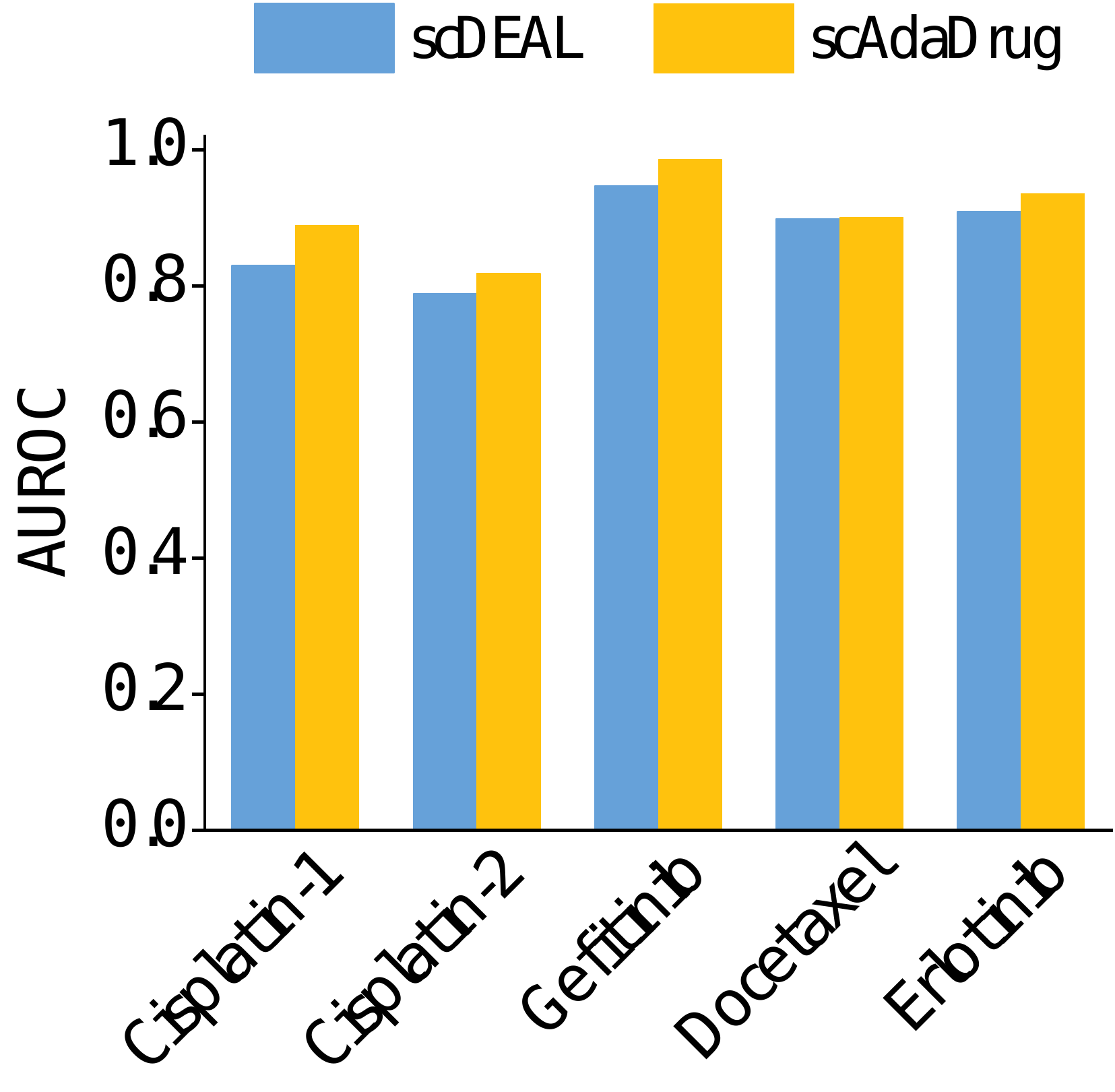} 
         \includegraphics[width=0.45\textwidth]{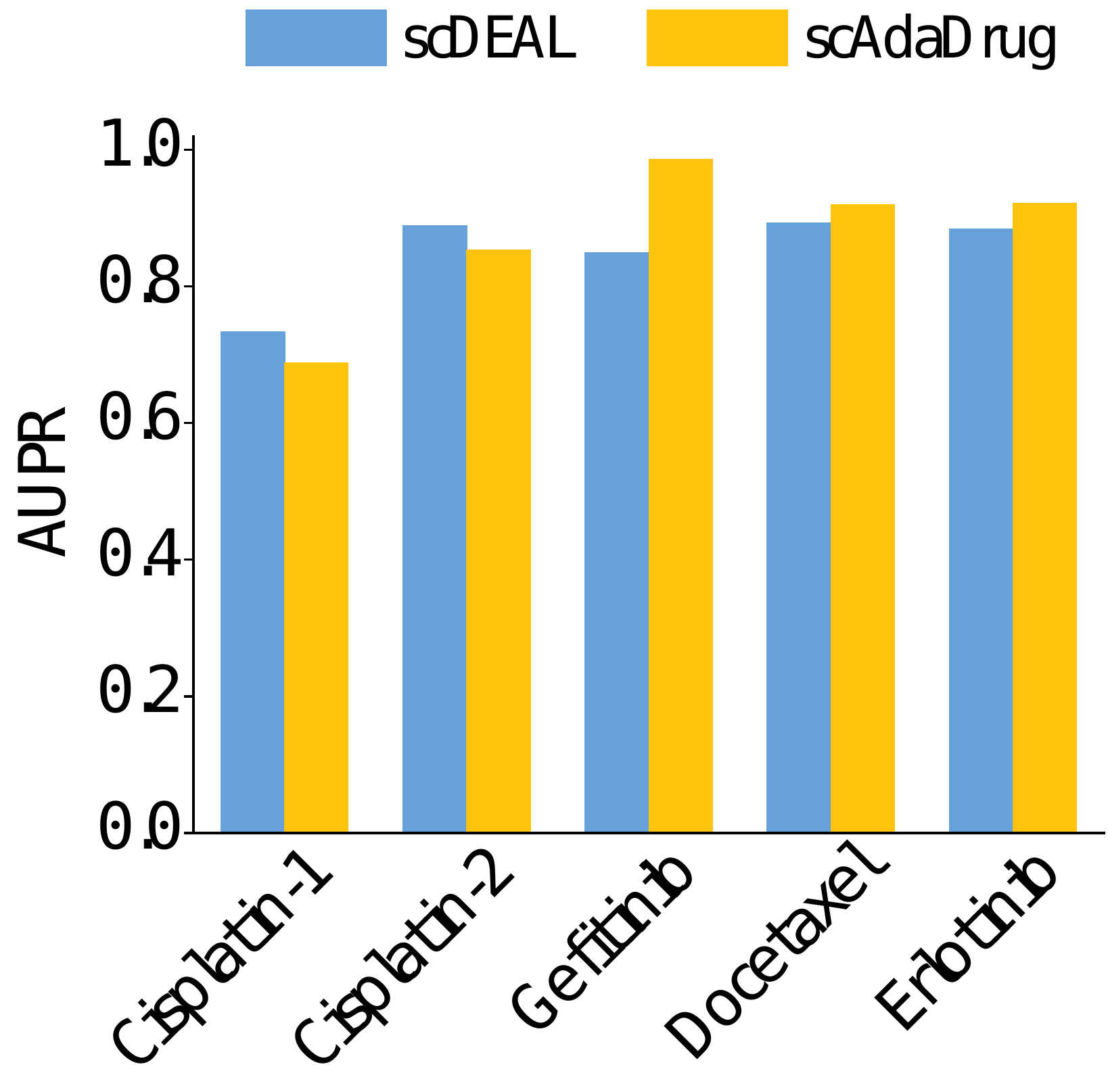} 
    \end{minipage}

    \caption{Comparison of AUROC and AUPR values between scAdaDrug and scDEAL on five single-cell drug sensitivity datasets.} \label{fig5:scDeal}
\end{figure}

Since scDEAL~\cite{chen2022deep} has been evaluated on these datasets, we conducted comparison experiments to demonstrate the performance of scAdaDrug. Similar to scDEAL, we employed SMOTE sampling to tackle the class imbalance problem. The experimental results indicated that scAdaDrug consistently exhibited high performance in predicting single-cell drug responses across all five datasets (Fig.~\ref{fig5:scDeal}). Particularly, the AUROC values of our method were dominantly superior to those of scDEAL on five datasets. Our method achieved an average AUROC value of 0.906 across five datasets, which is significantly higher than the 0.874 obtained by scDEAL (see details in TABLE S4).  Note that these datasets were complementary in the performance comparison of the methods, so only a method that works well on all datasets can demonstrate its superiority. These experimental results strongly validated the generalizability of our method to predict different drug-induced single-cell sensitivity of distinct cancer types.

% \begin{figure*}[h]
%     \centering
%     \begin{minipage}{0.45\textwidth}
%         \centering
%         \includegraphics[width=\textwidth]{fig7a_scdeal_auroc_comparison.pdf} 
         
%     \end{minipage}
%     \hfill
%     \begin{minipage}{0.45\textwidth}
%         \centering
%         \includegraphics[width=\textwidth]{fig7b_scdeal_aupr_comparison.pdf} 
        
%     \end{minipage}
%     \caption{Comparison of AUROC and AUPR values between scAdaDrug and scDEAL on five single-cell drug sensitivity datasets.} \label{fig7:scDeal}
% \end{figure*}

\subsection{Predicting patient-level drug response based on pathway activities}
To verify the extensible ability of our model, we went further to check whether our model can generalize from cell lines to patient clinical drug response. For this purpose, we retrieved the bulk RNA-seq and clinical response information from TCGA repository. Specifically, we selected the patients who underwent the treatment with one drug from a pool of six therapeutic agents, namely Cisplatin ($n$=420), Docetaxel ($n$=160), 5-fluorouracil ($n$=229), Gemcitabine ($n$=196), Paclitaxel ($n$=217), Sorafenib ($n$=37). Since the drugs were shared between GDSC cell lines and TCGA patients (see details in TABLE S5), we were able to build one model per drug to carry out performance evaluation. The clinical drug responses of TCGA patients were derived from clinical metadata, with complete response and partial response patients categorized as responders and patients with clinically progressive and stable diseases marked as non-responders. The common genes between cell lines and patients were selected to compute pathway activities, which were then taken as input to our model.  

\begin{figure}[htbp] % 使用H选项禁止图片浮动
  \begin{flushleft} % 左对齐
    \includegraphics[width=\textwidth]{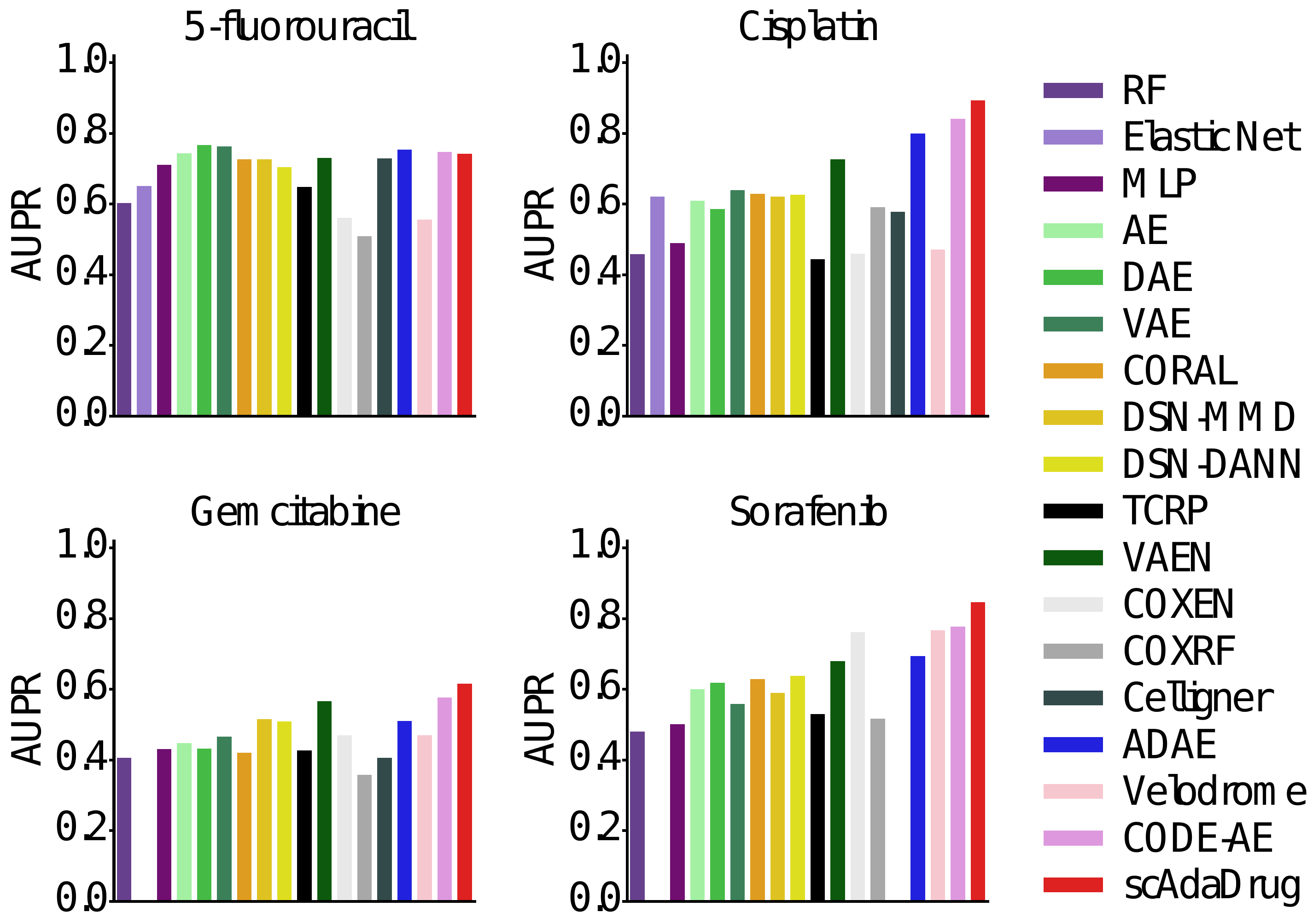} % 替换为你的图片文件名，并调整宽度
    \caption{Performance comparison of scAdaDrug and seventeen existing methods in predicting clinical patient responses to the treatments of four drugs, respectively.}
    \label{fig6_exp5_code-ae}
  \end{flushleft}
\end{figure}

Our model yields AUPR values of 0.893, 0.753, 0.742, 0.616, 0.689, and 0.846 in predicting the patient-level responses to six drug treatment, respectively. Furthermore, we compared our model to a variety of existing methods in predicting these TCGA patient response to therein four drugs (5-Fluorouracil, Cisplatin, Gemcitabine, Sorafenib), whose performance have been reported in CODE-AE~\cite{he2022context}. The experimental results showed that our model demonstrated better predictive performance than all previous methods (Fig.~\ref{fig6_exp5_code-ae}). Especially for Cisplatin and Sorafenib, our method achieved more significant advantage over all the comparative methods.

\subsection{Effectiveness of adaptive importance-aware feature}
To validate the effectiveness of different components included in our model, we conducted model ablation experiments on two single-cell drug sensitivity datasets, namely GSE149215 and GSE108383, with different configurations of upsampling algorithm and number of source domains. We designed three variants of our method as below:
\begin{itemize}
    \item w/o MDA: without multi-source domain adaptation. Only the source domains were used to train the autoencoder and predictor, which was then applied to predict single-cell drug sensitivity.
    \item w/o IND: without conditional independence constraint on the adaptively generated weights.
    \item w/o AWG: without the importance-aware weight generator. The source domains were assumed to contribute equally to domain transfer, and the features extracted by the autoencoder were aligned by adversarial domain adaptation without weight adjustment.\\
\end{itemize}
As shown in TABLE~\ref{tab:ablation_combined}, without the multi-source domain adaptation, the model performance declined most severely upon each combination of upsampling algorithm and source domain number. Also, the conditional independence constraint and importance-aware feature weighting contributed greatly to the performance.  Compared to the single-source domain model, our model exhibited great performance superiority. Interestingly, we observed that the removal of the conditional independence constraint and adaptive weight generator, the 2-source domain model outperformed 3-source domain model. We speculated that the increase in the number of source domains may cause more information redundancy, which posed difficulty for the encoder to extract essential feature. With these components integrated, our model yielded remarkable performance improvement with increasing number of source domains, indicating that these component collaborated to learn discriminative feature determining single-cell drug sensitivity.

\begin{table*}[h]
\centering
\caption{AUROC values achieved by ablated models on two drug-induced drug sensitivity at single-cell level}
\label{tab:ablation_combined}
\resizebox{\textwidth}{!}{%
\begin{tabular}{c|ccc|ccc|ccc|ccc} \hline
\multicolumn{1}{c|}{} & \multicolumn{6}{c|}{Weight sampling} & \multicolumn{6}{c}{SMOTE sampling} \\ \hline
Drug & \multicolumn{3}{c|}{Etoposide} & \multicolumn{3}{c|}{PLX4720} & \multicolumn{3}{c|}{Etoposide} & \multicolumn{3}{c}{PLX4720}\\ \hline
$N_{S_k}$ & 1 & 2 & 3 & 1 & 2 & 3 & 1 & 2 & 3 & 1 & 2 & 3 \\ \hline
w/o MDA & / & 0.614 & 0.634 & / & 0.765 & 0.712 & / & 0.606 & 0.633 & / & 0.814 & 0.738 \\
w/o IND & / & 0.656 & 0.652 & / & 0.789 & 0.752 & / & 0.674 & 0.669 & / & 0.819 & 0.763 \\
w/o AWG & / & 0.638 & 0.643 & / & 0.792 & 0.752 & / & 0.706 & 0.675 & / & 0.872 & 0.834 \\
scAdaDrug & 0.628 & \textbf{0.710} & \textbf{0.727} & 0.782 & \textbf{0.951} & \textbf{0.951} & 0.652 & \textbf{0.739} & \textbf{0.747} & 0.821 & \textbf{0.941} & \textbf{0.948} \\ \hline
\end{tabular}}
\end{table*}

\section{Discussion and Conclusion}
In this study, we introduced scAdaDrug, a new model for predicting drug sensitivity at single-cell level. By applying adaptively weighted features in multi-source domain adaptation, scAdaDrug exhibited exceptional performance in predicting the drug sensitivity of single cells, as well as patient samples. We believe that scAdaDrug could greatly contribute to our understanding of drug resistance mechanisms, highlighting its potential for precision medicine.

Compared to single-source domain model, our multi-source domain model notably enhanced the performance in predicting single-cell drug sensitivity across multiple types of cell lines. For patient-level drug responses, our model also surpassed the performance of single-source domain adaptation models like CODE-AE~\cite{he2022context}. More interestingly, we observed a positive correlation between the number of source domains and prediction accuracy. As illustrated in our experiment, the 3-source domain models consistently outperformed the 2-source domain setup in nearly all our tests. Therefore, we would attribute this advantage to our model's ability capture features determining drug sensitivity from multiple source domains, thanks to the innovative design of importance-aware feature weighting and conditional independence constraints.

Nevertheless, we observed that our method's performance is less than satisfactory when there is a substantial divergence in data distribution of target domains. Specifically, when the target domain data comes from cell lines, our method achieved average AUROC scores over 0.9 upon multiple drugs. However, the AUROC scores decline to between 0.7 and 0.9, when the target domains originates from PDX samples or patient tumor tissues. We postulate that this is primarily due to the greater tumor heterogeneity in PDX samples and patient tumor tissues compared to cell lines. The heterogeneity increases the data distribution disparity between the target and source domains, as well as the data drift within the target domain, which ultimately lead to reduced predictive accuracy. In addition, we encountered another challenge stemming from the low RNA abundance in individual cells, which frequently leads to some genes not being effectively captured by primer and their expression values appear as 0. In our practice, we filtered out genes with an excessive proportion of 0 values. However, the filtering process may introduce bias in the data distribution and compromised the model performance.

Finally, despite of superior performance in predicting single-cell drug sensitivity across various types of target-domain data, our current model cannot instantly adapt to new target domains. Instead, it requires to be retrained using the target-domain data of interest for domain adaptation, which somewhat limits the breadth of our model's applicability. Domain generalization, on the other hand, eliminates the need for target domain data during model development, allowing the trained model to be directly deployed on novel datasets. Looking ahead, we intend to focus on developing models rooted in domain generalization.

\section{Method}\label{sec2}

\subsection{Data source and preprocessing}\label{subsec1}
\subsubsection{Bulk drug sensitivity}
The bulk drug sensitivity data was retrieved from the Genomics of Drug Sensitivity in Cancer (GDSC) database~\cite{gretton2012kernel}, which provided drug sensitivities of 1074 cancer cell lines upon 226 drugs. Drug sensitivity was evaluated using half maximal inhibitory concentration (IC50) and area under the dose-response curve (AUC) measurements. We assigned drug response label to each cell line, by using the method similar to CODE-AE~\cite{he2022context}. For a drug of interest, we ranked the IC50 values of all cell lines tested against this drug, and then categorized them as sensitive or resistant. The binarization threshold was set as the average IC50 value across all cell lines tested against the drug. The sensitive cell lines were labeled as 1, while the resistant ones were labeled as 0. Together with the drug sensitivity data, we retrieved the matched bulk RNA-seq data (RMA-normalized basal expression profiles) of the cell lines from GDSC.

\subsubsection{Single-cell drug sensitivity}
For comprehensive performance evaluation, we collected the scRNA-seq data where cells were derived from different sources, including cell lines, patient-derived xenografts (PDX) models, and patient tumor tissues. The scRNA-seq data and single-cell drug response labels were obtained from the National Center for Biotechnology Information's (NCBI) Gene Expression Omnibus (GEO).

\subsubsection{Patient-level clinical drug response}
The patient transcriptional profiles was retrieved from TCGA repository (https://www.cancer.gov/tcga)~\cite{hutter2018cancer}, which included bulk RNA-seq data and clinical drug response information of the patients across various cancer types. The drug response labels of the TCGA patients were assigned using the similar method to a recent work~\cite{he2022context}. Specifically, the responders were those who had a partial or complete response to specific drug treatment, while the non-responders were those who had the progressive clinical disease or stable disease diagnosis. Only the patients received single-drug therapy through the entire duration of treatment were retained, while those treated by drug combinations were excluded. %It is noted that the patient drug response labels were not used in model training.

\subsubsection{Upsampling for minority class}
Since the unbalanced ratio of drug response labels in the source domains may affect the predictive performance, we leveraged upsampling methods to generate samples in the minority class. Two sampling algorithms, including weight-sampling~\cite{efraimidis2006weighted} and SMOTE-sampling~\cite{chawla2002smote}, were used in this study. The weight-sampling enhances the likelihood of minority categories being sampled assigning higher weights to minority categories and lower weights to majority categories~\cite{leevysurvey}, while SMOTE-sampling artificially creates synthetic samples of the minority class to create balanced dataset.

\subsection{scAdaDrug framework}
We proposed a novel method, referred to as scAdaDrug, which leverages the multi-source domain adaptation to transfer knowledge learned from bulk drug sensitivity of cell lines to predict single-cell drug response. In this study, we considered the bulk RNA-seq data of multiple cell lines of the same cancer type as the multi-source domains, and scRNA-seq data of the same cancer type as the target domain. Each cell line in the source domains has been assigned response label to specific drug. The scAdaDrug model consists of four components: autoencoder-based feature extractor, importance-aware weight generator, adversarial domain discriminator and drug sensitivity predictor (Fig.~\ref{fig:framework}). A shared autoencoder is employed to extract the features for both source and target domains. Specifically, we introduce an adaptive weight generator to produce an importance-aware vector for each source domain, which could capture element-wise relevance between source and target domains. Meanwhile, we required the adaptively generated weights were conditionally independent so as to eliminate information redundancy between multiple domain sources. Therefore, the adaptively generated weights allow for fine-grained and nonredundant knowledge transfer from source domains to target domain. Furthermore, we incorporated adversarial learning into our framework to learn domain-level invariant features related to drug response. The designed model is able to generalize well across different domains, thereby improving its predictive performance for drug sensitivity at the single-cell level. %This method enabling nuanced and effective transfer of knowledge across domains.

\subsubsection{Problem definition}
Suppose that there are $K$ source domains $\{S_1, S_2, \cdots, S_K\}$ and one target domain $T$. In the unsupervised multi-source domain adaptation scenario, source-domain samples are labeled and target-domain samples are unlabeled. Formally, we assume that there exist $N_{S_k}$ labeled samples $D_{S_k}=\left.\{(x_i^{S_k},y_i^{S_k})\}_{i=1}^{N_{S_k}}\right.$ in source domain $S_k$ ($k$=1,...,$K$), and $N_T$ unlabeled samples $D_T=\{x_i^T\}_{i=1}^{N_T}$ in target domain, where $x_i^{S_k}$ is the bulk transcriptional profile of $i$-th cell line from the $k$-th source domain $S_k$, $y_i^{S_k}$ referred to the corresponding label (sensitive or resistant) upon specific drug treatment, and $x_i^{T}$ represents the single-cell transcriptional profile of $i$-th cell from the target domain without label. An input sample consists of a tuple of source-domain samples and their true labels and a target-domain sample. For instance, $i$-th input is denoted by $(x_i^{S_1},..,x_i^{S_K}, y_i^{S_1},..y_i^{S_K}, x_i^{T})$.

\subsubsection{Autoencoder architecture}
We employ a shared autoencoder $E_{\theta}$ to extract features from expression profiles for both multiple source domains and target domain. Taking as input the source-domain sample $x_i^{S_k}$ and target-domain sample $x_i^{T}$, the encoder maps them to the embeddings $h_i^{S_k}=E_{\theta}(x_i^{S_k})$ and $h_i^T=E_{\theta}(x_i^{T})$ in the latent space. The decoder is defined as $G_{\varphi}$ that converts the embedding to output space. Our encoder/decoder networks are fully or densely connected neural networks with rectified linear unit (ReLU) activation function, $\theta$ and $\varphi$ are the learnable parameters of the encoder and decoder. We require that the encoder converts the input expression profiles into low-dimensional but informative representations in the latent space, while the decoder tries to reconstruct the input samples from its as much as possible. Instead of directly utilizing the representations generated by the encoder to reconstruct input samples, we employ importance-weighted representations for the reconstruction process. Suppose that $z_i^{S_k}$ and $z_i^T$ represent the importance-weighted representations for $h_i^{S_k}$ and $h_i^T$ (see Section~\ref{sec:generator} for details), we optimize the parameters by minimizing the reconstruction loss as below:

\begin{equation}
\mathcal{L}_{reco}(\theta,\varphi) =\sum_{k=1}^{K}\left(\frac{1}{N_{S_k}}\sum_{i=1}^{N_{S_k}}\|G_{\varphi}(z_i^{S_k})-x_i^{S_k}\|^2\right)+\frac{1}{N_T}\sum_{j=1}^{N_T}(\|G_{\varphi}(z_i^{T})-x_i^{T}\|^2) 
\end{equation}

\subsubsection{Importance-aware weight generator}\label{sec:generator}
Most multi-source domain adaptation methods assume that the source domains have uniform importance in transferring knowledge, neglecting the inherent difference in their contribution to target domain tasks. Inspired by MDDA~\cite{zhao2020multi}, we realize that different source domains have varying levels of importance in the context of domain transfer. To capture the correlation of features between source and target domain in dimension level, we introduce an adaptive weight generator to produce importance-aware vectors, which assigns element-wise weights to the embeddings of the source and target samples. The generator is actually a two-layer fully-connected network, which takes as input the embeddings of source and target samples and automatically learns to produce the weight vector representing the importance of different dimensions. Formally, let $F_{\vartheta}$ represent the function of generator parameterized by $\vartheta$, we have
\begin{equation}
w_{i}^{S_k}=F_{\vartheta}(|h_i^T-h_i^{S_k}|)
\end{equation}
in which $w_{i}^{S_k}$ is the generated weight for the sample $i$ of source domain $S_k$ paired to target domain sample $j$ in a mini-batch, and $|\cdot|$ represents the abs function. The subtract of $h_i^T$ by $h_i^{S_k}$ represents the similarity between the source and target samples. After the generation of $w_{i}^{S_k}$, we weight the embedding $h_i^{S_k}$ as below:
\begin{equation}
z_i^{S_k}=h_i^{S_k}\odot w_{i}^{S_k}
\end{equation}
where $\odot$ represents the element-wise product. For the target sample, we compute the mean weight vector regarding $K$ source domains as $\bar{w}_i=\frac{1}{K}\sum_{k=1}^K w_{i}^{S_k}$, and then apply it to the target sample embedding by element-wise product.

Meanwhile, to avoid information redundancy in transferring knowledge from source domain to target domain, we also impose conditional independence constraint on the adaptively generated weights. Specifically, we require that the generated weight vectors should be mutually orthogonal. Denote the generated weight matrix by $W=\{{w}_i^{S_1}, {w}_i^{S_2},...,{w}_i^{S_K}\}$, we define the following loss function:
\begin{equation}
    \mathcal{L}_{ind}(\omega)=\frac{1}{2}||W W^T-I||_F^2
\end{equation}
where $I$ is the identity matrix, and $F$ denotes the Frobenius norm. The minimization of $\mathcal{L}_{ind}$ make the non-diagonal elements of $W W^T$ close to 0, thereby enforcing the generated weight vectors to be mutually independent. 

It is worth noting that our importance-aware weight generator differs from PFSA~\cite{fu2021partial} in at least two aspects. First, instead of using uniform weight vector for all source domain features, our method assigns unique feature weighting vector to each individual source domain sample. This approach implicates dimension-level weighting scheme that effectively underscores the significant dimensions within the latent feature space. Second, we introduce the conditional independence constraint on the adaptively generated weights to minimize redundant information among source domain features, thus allowing us to establish the explicit connection for transferring knowledge from source to target domain.

\subsubsection{Adversarial domain adaptation}
For adversarial domain adaptation, a domain discriminator $D_{\psi}$ is introduced to distinguish the domain origin (source or target) of the features. The feature extractor aims to produce features that confuses the domain discriminator, while the domain discriminator strives to correctly identify the domain of the features. Through adversarial training, the feature extractor can learn to produce domain-invariant representations, while still preserving sufficient discriminative information for the main task. For this purpose, we assign a domain label of 1 to the source domain features and a label of 0 to target domain features. Denote by $L_{adv}$ the domain discrimination loss, the domain discriminator is trained to minimize the binary-cross entropy (BCE) loss as below
\begin{equation}
\mathcal{L}_{adv}(\psi,\theta)=\min_{D_{\psi}}\max_{E_{\theta}}-\sum_{k=1}^K \sum_{i=1}^{M} (y_D\log D(z_i^{S_k})+(1-y_D)\log(1-D(z_i^{S_k})))
\end{equation}
where $y_D$ is the actual domain label regarding the sample $z_i^{S_k}$, $D(z_i^{S_k})$ represents the predicted domain label by the domain discriminator, and $M$ is the total number of samples from both source and target domains included in a mini-batch.

\subsubsection{Drug response predictor}
The bulk drug response labels of cell lines to specific drug exposure is used to train the predictor in the source domains. The predictor $P$ takes as input the weighted features of cell lines to predict their labels. Therefore, the classification loss $\mathcal{L}_{cls}$ is defined as below:
\begin{equation}
\mathcal{L}_{cls}(\phi)=\sum_{k=1}^K(\sum_{i=1}^{N_{S_k}}\|y_i^{S_k}-P_{\phi}(z_i^{S_k})\|^2)
\end{equation}
in which $\phi$ denotes the parameters of the predictor. Once the drug response predictor has been trained on source domain labels, the domain-invariant features learned by adversarial domain adaptation allow to predict drug responses in single cells (the target domain) effectively. 

Finally, the total loss function is defined as the sum of the four loss terms as below:
\begin{equation}
    \mathcal{L}(\theta,\varphi,\omega,\psi,\phi)=\mathcal{L}_{reco}+\mathcal{L}_{ind}+\mathcal{L}_{adv}+\mathcal{L}_{cls}
\end{equation}
in which the parameters are optimized during the minimization of the loss function.

Once the model training is finished, the predictor is used to predict drug response for target-domain samples.

\backmatter

% \bmhead{Supplementary information}

% If your article has accompanying supplementary file/s please state so here. 

% Authors reporting data from electrophoretic gels and blots should supply the full unprocessed scans for key as part of their Supplementary information. This may be requested by the editorial team/s if it is missing.

% Please refer to Journal-level guidance for any specific requirements.

% \bmhead{Acknowledgements}

% Acknowledgements are not compulsory. Where included they should be brief. Grant or contribution numbers may be acknowledged.

% Please refer to Journal-level guidance for any specific requirements.

\section*{Declarations}
\begin{itemize}
\item Funding \\
This work was supported by National Natural Science Foundation of China (No. 62072058, No. 62372229), Natural Science Foundation of Jiangsu Province (No. BK20231271).
\item Conflict of interest\\
The authors declare no competing interests.
\item Ethics approval and consent to participate \\
Not applicable.
%\item Consent for publication
\item Data availability \\
GDSC is publicly available through the website (\url{https://www.cancerrxgene.org/}). Drug response annotation, including half maximal inhibitory concentration (IC50) and area under the dose-response curve (AUC), are available through the page \url{https://www.cancerrxgene.org/downloads/bulk_download}. Gene expression data (RMA-normalized basal expression profiles) for cell lines can be accessed on GDSC (\url{https://www.cancerrxgene.org/gdsc1000/GDSC1000_WebResources/Home.html}). The eight scRNA-seq data used in this study are available in the Gene Expression Omnibus (\url{https://www.ncbi.nlm. nih.gov/geo/}) under accession numbers GSE149215, GSE108383, GSE175716, GSE223779, GSE112274, GSE117872, GSE140440 and GSE149383. Detailed descriptions of scRNA-seq datasets used in this study can be found in Supplementary Data. The bulk RNA-seq and clinical information of TCGA patients are available from the NIH genomic data commons (\url{https://portal.gdc.cancer.gov}). The Gene sets for calculating pathway activity are available at \url{https://www.gsea-msigdb.org/gsea/msigdb}. 

\item Code availability \\
The source codes and datasets used in this study are available at: \url{https://github.com/hliulab/scAdaDrug}.
\item Author contribution \\
H.L. and  W.D. conceptualized the idea. H.L. and W.D. implemented the model. W.D. collected the data and conducted the experiments. W.D. plotted figures. H.L. and W.D. prepared the manuscript. H.L. and J.L. revised the manuscript. J.L. supervised the research.
\end{itemize}

%\noindent
%If any of the sections are not relevant to your manuscript, please include the heading and write `Not applicable' for that section. 



\bibliography{reference}% common bib file

\newpage

\end{document}